\documentclass[letterpaper, 10 pt, conference]{ieeeconf}  

\IEEEoverridecommandlockouts                              

\usepackage{graphics} 
\usepackage{epstopdf}
\usepackage{epsfig} 
\usepackage{subfigure}
\usepackage{amsmath} 
\usepackage{multirow}
\usepackage{threeparttable}
\usepackage{enumerate}
\usepackage{array}
\usepackage{url}

\title{\LARGE \bf
DeepVIO: Self-supervised Deep Learning of Monocular Visual Inertial Odometry using 3D Geometric Constraints
}

\author{Liming Han$^{1}$, Yimin Lin$^{1}$, Guoguang Du$^{1}$, Shiguo Lian$^{1}$
\thanks{$^{1}$Liming Han, Yimin Lin*, Guoguang Du, Shiguo Lian are all with the AI Department, CloudMinds Technologies Inc., Beijing, 100102, China
        {\tt\small {liming.han, george.du, scott.lian}@cloudminds.com, *linyimin2012@hotmail.com}}%
}

\begin{document}

\maketitle
\thispagestyle{empty}
\pagestyle{empty}

\begin{abstract}
This paper presents an self-supervised deep learning network for monocular visual inertial odometry (named DeepVIO). DeepVIO provides absolute trajectory estimation by directly merging 2D optical flow feature (OFF) and Inertial Measurement Unit (IMU) data. Specifically, it firstly estimates the depth and dense 3D point cloud of each scene by using stereo sequences, and then obtains 3D geometric constraints including 3D optical flow and 6-DoF pose as supervisory signals. Note that such 3D optical flow shows robustness and accuracy to dynamic objects and textureless environments. In DeepVIO training, 2D optical flow network is constrained by the projection of its corresponding 3D optical flow, and LSTM-style IMU preintegration network and the fusion network are learned by minimizing the loss functions from ego-motion constraints. Furthermore, we employ an IMU status update scheme to improve IMU pose estimation through updating the additional gyroscope and accelerometer bias. The experimental results on KITTI and EuRoC datasets show that DeepVIO outperforms state-of-the-art learning based methods in terms of accuracy and data adaptability. Compared to the traditional methods, DeepVIO reduces the impacts of inaccurate Camera-IMU calibrations, unsynchronized and missing data.
\end{abstract}

\section{INTRODUCTION}\label{sec:Intro}

6-DoF motion estimation is one of the key challenges in the fields such as robotics and autonomous driving. Because of low-cost, and easy hardware setup,  camera-based solutions \cite{fraundorfer2012VO}  have drawn a large attention by the community. Therefore, in the last decade, impressive results have been demonstrated in the contexts of Visual Odometry (VO) and Visual Simultaneous Localization and Mapping (VSLAM) \cite{cadena2016VSLAM}. For example, the direct-based representative DSO \cite{engel2016DSO} and the feature-based representative ORB-SLAM \cite{mur2017orbslam2}  both achieve very high localization accuracy in the large-scale environment, and real-time performance with commercial CPUs. However, they still face some challenging issues when they are deployed in non-textured environments, serious image blur or under extreme lighting conditions. Recently, many Visual Inertial Odometry (VIO) systems \cite{sun2018sMSCKF, Mur2016VIORB, Tong2017VINS} are proposed to eliminate these limitations, which combine measurements from an inertial measurement unit (IMU) with a camera to improve motion tracking performance. However, the current VIO systems heavily rely on manual interference to analyze failure cases and refine localization results. Furthermore, all these VIO systems require careful parameter tuning procedures for the specific environment they have to work in.

\begin{figure}[!htbp]
\centering
\includegraphics[width=0.45\textwidth]{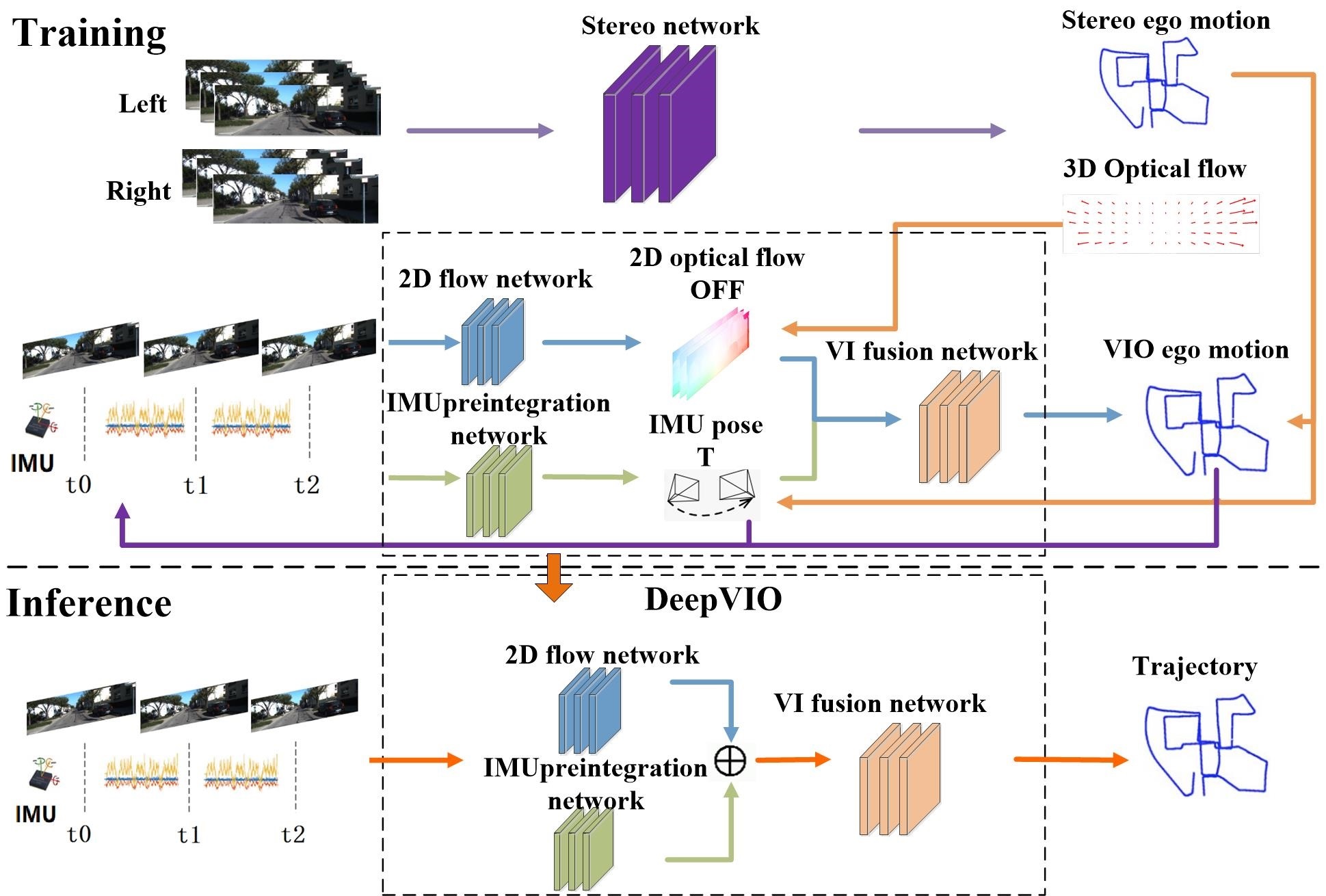}
\caption{The pipeline of the DeepVIO. Novel 3D optical flow and stereo ego motion are used as 3D geometric constraints to supervise 2D optical flow learned from 2D flow network, ego-motions estimated from the IMU preintegration network and VI fusion network, the state of the IMU is updated when it receives the feedback from VI fusion network. More details can be found in Sec. III.}
\label{figure:figure1}
\end{figure}

In recent years, deep learning based VO has drawn significant attentions due to its potentials in learning capability and the robustness to camera parameters and challenging environments \cite{Wang2017DeepVO, wang2018deepvo}. These data-driven VO methods have successfully learned new feature representations from images that are used to further improve the motion estimation. Inspired by this, a deep learning based VIO \cite{Clark2017VINet} is also demonstrated to fuse the estimated pose from deep Convolutional Neural Networks (CNNs) and the IMU data with the Long Short-Term Memory (LSTM) networks. Although supervised deep learning methods have achieved state-of-the-art results on motion estimation, it is impractical to gather large amounts of ground truth data. Moreover, the trained models usually do not generalize well to unseen scenes without fine-tuning on sufficient ground truth data. These limitations suggest some methods \cite{yin2018geonet, shamwell2018vision} to look for various unsupervised learning VO/VIO schemes, and consequently learn a mapping from pixels to flow, depth and camera motion without trajectory ground truth. Nevertheless, most unsupervised methods learn from photometric and temporal consistency between consecutive frames in a monocular video, which are prone to scale-ambiguity. Meanwhile, they try to regress the pose but probably result in high drift since there are not enough data covering various moving speed of single camera. Recently, 2D optical flow is widely used as self-supervisory signals to learn unsupervised ego-motion system, but it has aperture problem due to the missing structure in local parts of the single camera.  Moreover, it cannot explicitly handle dynamic objects and textureless environments from consecutive frames that inevitably results in inaccurate flow predictions.

To overcome these limitations, a stereo camera  based approach is proposed in this paper as shown in Fig.~\ref{figure:figure1}. This approach not only provides synthetic 2D optical flow as supervision from its precise 3D optical flow, but also solves the scale and speed ambiguity using its absolute 6-DoF pose. Additionally, an IMU status update scheme is designed to improve IMU pose estimation. Some experimental results will be given to show that, the proposed DeepVIO is able to estimate VIO in a real world scale from monocular camera and IMU sequences. In summary, our main contributions are as follows:

(1) It is the first time to present an self-supervised end-to-end monocular VIO network with the supervisory signals obtained from the stereo sequences.

(2) It takes 3D geometric constraints including 3D optical flow and 6-DoF pose to penalize inconsistencies in the estimated 2D OFF, IMU pose and VIO trajectory.

(3) We update additional bias for IMU using the pose feedback from FC-fusion network similar to traditional  tightly-coupled VIO methods.

(4) Our experimental results achieve good performance in terms of accuracy and data adaptability compared to state-of-the-art learning based VO and VIO systems on KITTI and EuRoC datasets.

\section{RELATED WORK}\label{sec:review}

\subsection{Traditional Methods}\label{sec:review:slam}

VIO fuses raw camera and IMU data in a single pose estimator and leads to more robust and higher accuracy even in complex and dynamic environments. In the past several decades, most tightly-coupled VIO systems can be divided into filtering-based and optimization-based approaches. Filtering-based representative is MSCKF\cite{mourikis2007MSCKF}, which combines geometric constraints with IMU measurements in a multi-state constraint Extended Kalman Filter (EKF). The method has low computational complexity and is capable of very accurate pose estimation in large-scale real environments. ROVIO\cite{bloesch2015Rovio} is another popular filtering based VIO. It uses an EKF to fuse the IMU data and intensity errors in the update step. On the other hand, optimization-based representative is OKVIS\cite{leutenegger2013OKVIS}, which is also called a keyframe-based VIO system. VINS\cite{Tong2017VINS} is a tightly-coupled, nonlinear optimization-based method, which is used to obtain highly accurate odometry by fusing preintegrated IMU measurements and feature observations.

\subsection{Supervised or Semi-supervised Learning Methods}\label{sec:review:slam}

Supervised learning based approaches have recently achieved great advances in motion estimation \cite{li2018ongoing}. Avoiding the hand-crafted features extraction used in previous methods, the deep neural networks  have successfully learned good feature representation directly from a lot of images. \cite{rambach2016learning} presents a LSTM to learn the relation between camera poses and inertial sensor measurements. This estimation is then appropriately combined with the output of a visual tracking system using a linear Kalman Filter to provide a robust final pose estimation. In addition, \cite{Clark2017VINet} fuses the estimated pose from \cite{Wang2017DeepVO} and the IMU data with the LSTM. The fusion network is trained jointly in an end-to-end way, and the proposed fusion system demonstrates comparable performance with traditional sensor fusion methods. However, these supervised learning approaches need large amounts of ground truth data for training, which is expensive to obtain in practice.

\subsection{Unsupervised Learning Methods}\label{sec:review:slam}

In recent years, large progress has been achieved in the development of depth estimation with unsupervised learning methods \cite{garg2016unsupervised, godard2017unsupervised}. More and more unsupervised paradigms for ego-motion estimation have been proposed to exploit brightness constancy and spatial smoothness to train depth or flow models. All the networks are jointly optimized during training, and then they can be applied independently at test time. For instance, \cite{zhou2017unsupervised} learns depth and ego-motion from monocular video in an unsupervised way. The CNNs is trained by a photometric reconstruction loss, which was obtained by warping nearby views to the target using the computed depth and pose. In addition, \cite{casser2018depth} addresses unsupervised learning of scene depth, robot ego-motion and object motions where the supervision is provided by geometric structure of monocular videos as input. Furthermore, many recent efforts \cite{yin2018geonet, zou2018dfnet} explore the geometric relationships between depth, camera motion, and flow for unsupervised learning of depth and flow estimation models. However, all the methods focus on monocular setup, and an extension to the above frameworks is to use stereo pairs for training. The main advantage is that this avoids issue with the depth-speed ambiguity that exists in monocular 3D reconstruction. \cite{li2018undeepvo} and \cite{zhan2018unsupervised} both use stereo pairs to define loss function for training the networks from spatial and temporal dense information. At test time their frameworks are able to estimate depth and two-view odometry from a monocular sequence.

To the best of our knowledge, \cite{shamwell2018vision} is the first unsupervised VIO system. The network learns to integrate IMU measurements and generate trajectories which are then corrected online according to the Jacobians of scaled image projection errors with respect to a spatial grid of pixel coordinates. However, it needs depth sensor, like RGBD or Lidar sensor, as input in order to achieve absolute scale recovery. Different from this, this paper proposes an self-supervised VIO system using a low-cost and easy-to-use stereo camera instead.

\begin{figure*}[t]
\centering
\includegraphics[width=\textwidth]{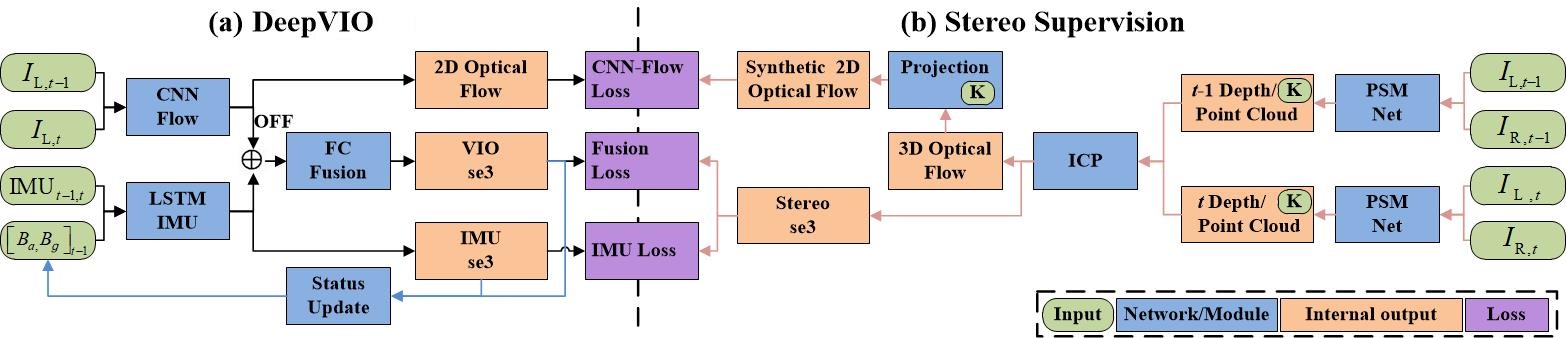}
\caption{Illustration of our proposed framework in training and inferring phase. DeepVIO consists of CNN-Flow, LSTM-IMU and FC-Fusion, which is trained by the supervisory signals obtained from 3D geometric constraints of stereo sequences (e.g., CNN-Flow Loss, IMU Loss and Fusion Loss), it updates IMU status using the pose feedback from FC-fusion network.}
\label{figure:figure2}

\end{figure*}

\section{METHOD}\label{sec:review}

The proposed DeepVIO mainly consists of CNN-Flow, LSTM-IMU and FC-Fusion networks, which jointly compute continuous trajectories from monocular images and IMU data as shown in Fig.~\ref{figure:figure2} (a). More specifically, the OFF and 2D optical flow are calculated by the CNN-Flow network through the monocular sequences (e.g., $I_{{\rm L},t-1}$ and $I_{{\rm L},t}$). Meanwhile, the relative 6-DoF pose (e.g., IMU-se3) between the adjacent two frames is calculated by IMU preintegration network (LSTM-IMU) through the IMU data (e.g., $IMU_{t-1,t}$) and status $S[Ba,Bg]$ (bias of accelerometer and gyroscope). Finally, the concatenated features from OFF and IMU-se3 are fed into the FC-Fusion network to calculate the trajectory of the monocular camera (e.g., VIO-se3). Note that, VIO-se3 and IMU-se3 give IMU status corrections as a feedback through the status update module.

Here, we employ stereo sequences as supervision for learning DeepVIO in Fig.~\ref{figure:figure2} (b). In particular, the pretrained depth network (e.g., PSMNet) is firstly applied to estimate dense depth images from left and right images at time $t-1$ (e.g., $t-1$ Depth) and $t$ (e.g., $t$ Depth), respectively. After that, we can recover 3D point clouds from each pair of left image and its depth. Next, the 6-DoF relative pose (e.g., Stereo-se3) and 3D optical flow are calculated from the two point clouds (e.g., $t-1$-th and $t$-th point clouds) via the well-known Iterative Closest Point (ICP) method \cite{yang2016goICP}. Moreover, we synthetize a 2D optical flow by projecting the 3D optical flow into its view. During DeepVIO training, the 3D geometric constraints are:

\begin{enumerate}[$\bullet$]
\item Optical Flow Loss. It provides pixel-level constraints for training CNN-Flow network from minimizing the differences between synthetic 2D optical flow and inferred 2D optional flow.
\item IMU Loss. LSTM-IMU is optimized by minimizing the differences between its inferred IMU-se3 and the supervision Stereo-se3.
\item Fusion Loss. Similar to IMU Loss, the output of FC-Fusion network (e.g. VIO-se3) is also constrained by Stereo-se3.
\end{enumerate}

More details are presented in the following sections.

\subsection{Stereo Network as Supervision}\label{sec:review:slam}

Fig.~\ref{figure:figure2} (b) shows how to generate 3D geometric constraints including 3D optical flow and Stereo-se3 from stereo sequences.

\subsubsection{Depth and Point Cloud}\label{sec:review:slam}

\begin{figure}[!htbp]
\centering
\subfigure[Left image]{%
\includegraphics[width=0.2\textwidth]{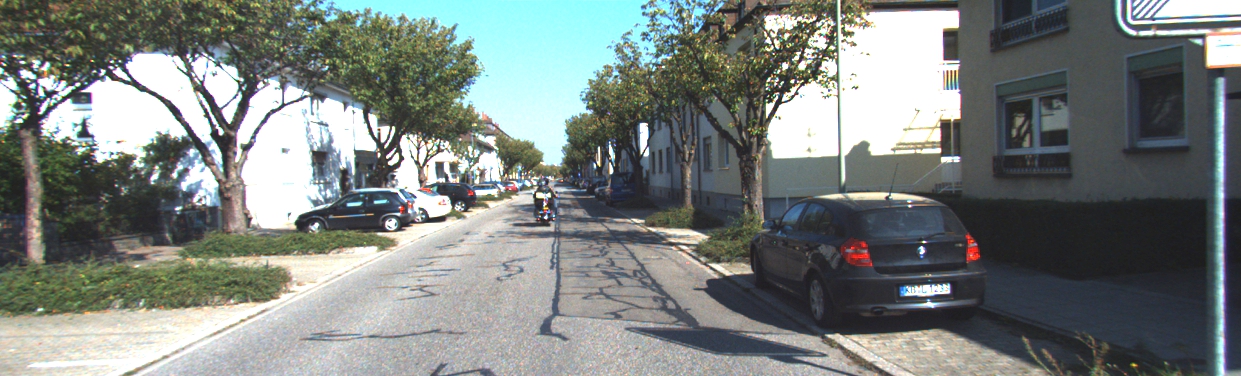}
\label{fig:subfigure1}
}
\subfigure[Right image]{%
\includegraphics[width=0.2\textwidth]{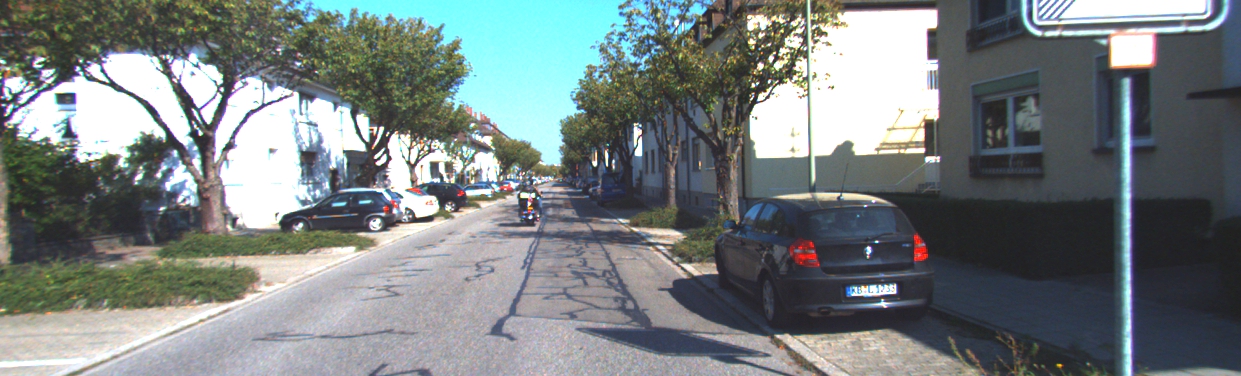}
\label{fig:subfigure2}
}
\quad
\subfigure[Disparity map]{%
\includegraphics[width=0.2\textwidth]{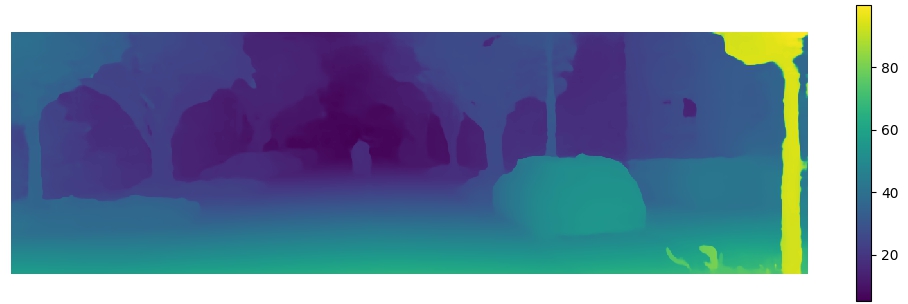}
\label{fig:subfigure3}
}
\subfigure[Point cloud]{%
\includegraphics[width=0.2\textwidth]{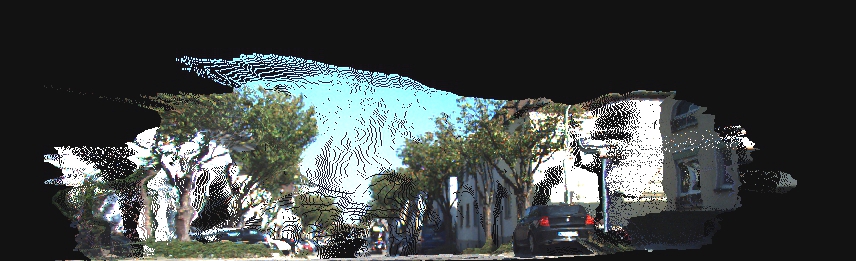}
\label{fig:subfigure4}
}
\quad
\caption{Illustration of  disparity map and point cloud obtained from stereo images. Where  (a) is the left image, (b) is the right image, (c) is the disparity map and (d) is the 3D point cloud, respectively.}
\label{fig:figure3}
\end{figure}

In most stereo systems, the disparity map is calculated through traditional stereo matching algorithms \cite{scharstein2002taxonomy} or directly estimated by an end-to-end CNNs \cite{chang2018pyramid}. In this paper, we chose the state-of-the-art PSMNet \cite{chang2018pyramid} since it provides the accurate and dense disparity map. Note that, PSMNet¡¯s output scene disparity map $q_{\rm L}=(x_{\rm L},y_{\rm L},x_{\rm L}-x_{\rm R})$, where the coordinates $(x_{\rm L},y_{\rm L})$ give the position of a point $P$ in left image plane, and its matching point $(x_{\rm R},y_{\rm R})$ in right image where $y_{\rm L}=y_{\rm R}$ in such rectified stereo images. Then the depth $d_{\rm L}$( see Fig.~\ref{fig:figure3}(c) ) are calculated as,

\begin{equation}
d_{\rm L} = \frac{fb}{q_{\rm L}} = \frac{fb}{x_{\rm L}-x_{\rm R}}
\end{equation}

where $b$ is the baseline and $f$ is the camera focal length. Once we get the depth $d_{\rm L}$, the 3D point cloud $\mathbf{c}$ is reconstructed as£º

\begin{equation}
\mathbf{c}=\mathbf{K}^{-1} d_{\rm L}[x_{\rm L},y_{\rm L},1]^{T}
\end{equation}

where $\mathbf{K}$ is the intrinsic parameter of left camera \cite{Cheng2018Learning}. Fig.~\ref{fig:figure3}(d) shows the 3D point cloud converted from each pixel of depth map with absolute scales.

\subsubsection{Stereo-se3 with ICP}\label{sec:review:slam}

In order to reduce the estimation error from depth prediction of PSMNet, only a part of points whose depth values satisfy $d_1<d_{\rm L}<d_2$ are selected to reconstruct the 3D point cloud, where $d_1$ and $d_2$ are the nearest and farthest depth threshold. For instance, the depth estimated from the sky region has a large error and is necessary to eliminate in our case. Then we generate two adjacent 3D point clouds $\mathbf{c_{t-1}}$ and $\mathbf{c_t}$ follows the Eq.(2). Then we implement the traditional ICP algorithm as,

\begin{equation}
\left\{I(\mathbf{c_{t-1}},\mathbf{c_t}),\mathbf{R},t\right\}={\rm ICP}(\mathbf{c_{t-1}},\mathbf{c_t})
\end{equation}

where $I(\mathbf{c_{t-1}},\mathbf{c_t})$ is the index of matched 3D points, and $\mathbf{R}$,$t$ is rotation and translation.  Obviously, they have the following properties as.

\begin{equation}
{\rm \mathbf{T}}=\left\{\left(\begin{array}{cc}
     \mathbf{R} & \mathbf{t}\\
     0    & 1\\
\end{array}\right)|\mathbf{R} \in SO(3),\mathbf{t} \in R^3\right\}
\end{equation}

thus Stereo-se3 $[\mathbf{\omega},\mathbf{\upsilon}]$ is a 6-dimension vector ($R^6$) that is calculated as,
\begin{equation}
se3={\rm log}\left\{\mathbf{R},\mathbf{t} \right\}=\left\{\left(\begin{array}{cc}
     \mathbf{\omega}    & \mathbf{\upsilon}\\
     0    & 0\\
\end{array}\right)|\mathbf{\omega} \in so(3),\mathbf{\upsilon} \in R^3\right\}
\end{equation}

\subsubsection{3D Optical Flow}\label{sec:review:slam}

Finally, the 3D optical flow in the scene can be calculated by the corresponding 3D points in two point clouds as,

\begin{equation}
(v_X,v_Y,v_Z)=\Delta I(\mathbf{c}_{t-1},\mathbf{c}_t)=\mathbf{c}_{t-1}-\mathbf{c}_t
\end{equation}

where $\mathbf{v}_{\rm 3D}=(v_X,v_Y,v_Z)$ is the 3D optical flow in the real world. In order to visualize our 3D optical flow , we reproject the spare $\mathbf{v}_{\rm 3D}$ to left and right views as,
\begin{equation} \label{eqn2}
  \begin{split}
  (v_x,v_y,1)_{\rm L}&=\frac{\mathbf{K} \mathbf{v}_{\rm 3D}}{d_L(x,y)}\\
  (v_x,v_y,1)_{\rm R}&=\frac{\mathbf{K} (\mathbf{R} \mathbf{v}_{\rm 3D}+t)}{d_L(x,y)}
  \end{split}
\end{equation}

\begin{figure}[!htbp]
\centering
\subfigure[]{%
\includegraphics[width=0.2\textwidth]{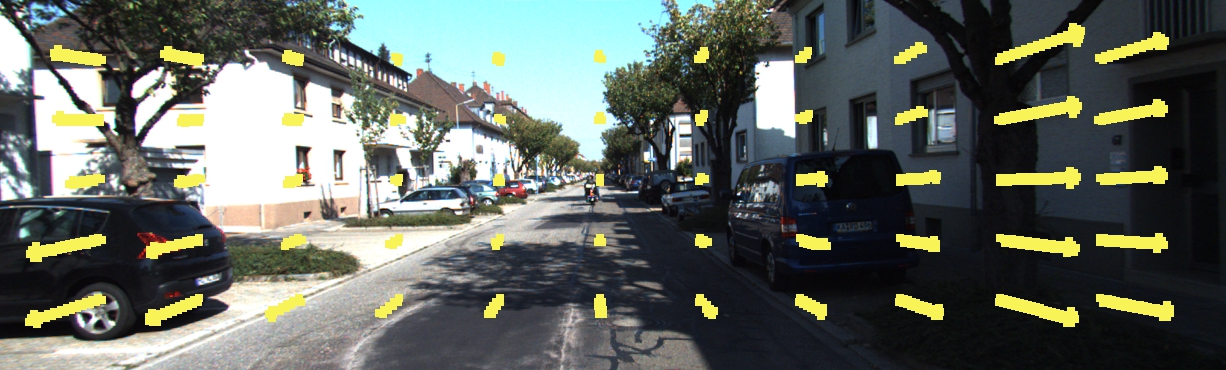}
\label{fig:subfigure1}
}
\subfigure[]{%
\includegraphics[width=0.2\textwidth]{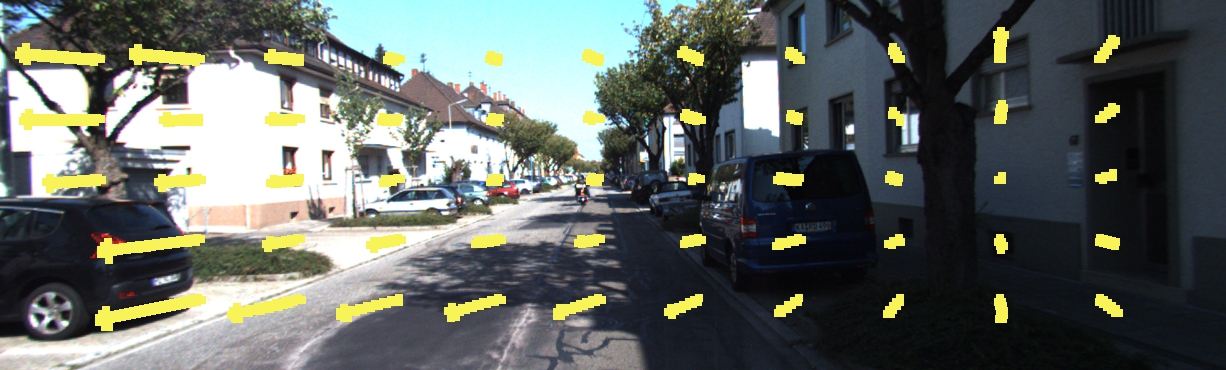}
\label{fig:subfigure2}
}
\quad
\subfigure[]{%
\includegraphics[width=0.2\textwidth]{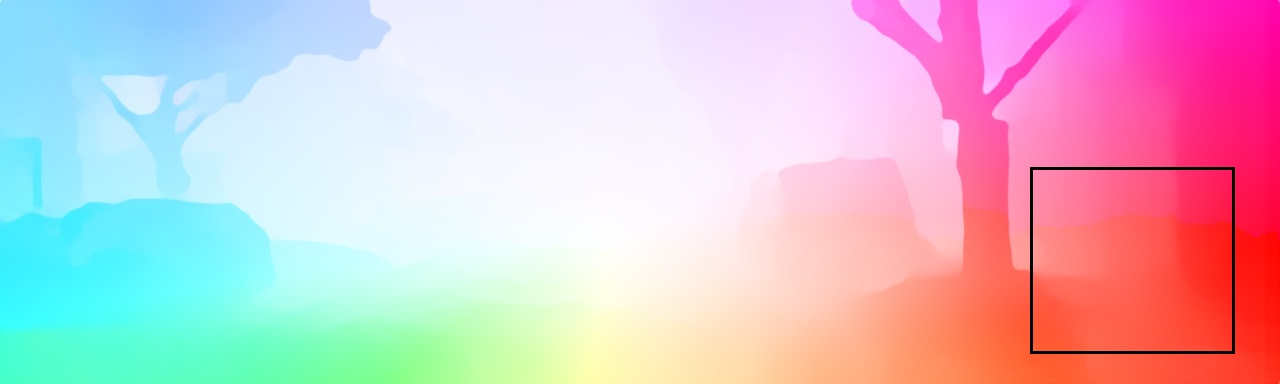}
\label{fig:subfigure3}
}
\subfigure[]{%
\includegraphics[width=0.2\textwidth]{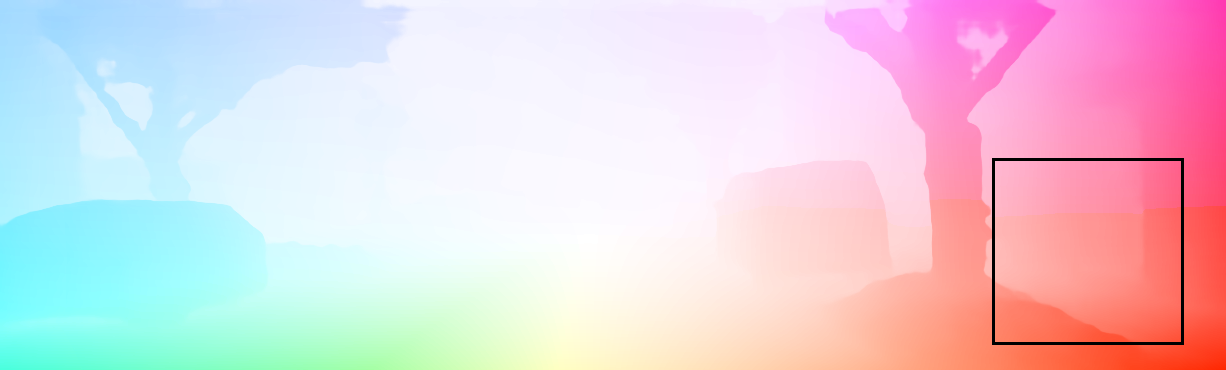}
\label{fig:subfigure4}
}
\quad
\caption[]{Illustration of 3D optical flow from (a) left and (b) right views, respectively (shown as yellow arrows), (c) synthetic 2D optical flow calculated form our 3D optical flow and (d) 2D optical flow obtained from FlownetC. It can be seen from (c) and (d) that our 3D optical flow represents the real world more accurate.}
\label{figure:figure4}
\end{figure}

Thus we can show them in left and right images (see arrows in Fig.~\ref{figure:figure4}(a) and (b)) using the Eq.7.

\subsubsection{Synthetic 2D Optical Flow}\label{sec:review:slam}

We obtain a synthetic 2D optical flow named as $\mathbf{v}_{\rm 2D}=(v_x,v_y)$ by projecting every pixels of 3D optical flow to the left image using Eq.(7), as shown in Fig.~\ref{figure:figure4}(c). In the KITTI dataset, true value of the optical flow is provided using Velodyne HDL-64E, but it is a sparse optical flow. In contrast, our synthetic 2D optical flow is a dense and each pixel has a true corresponding optical value.

Intuitively, traditional monocular 2D optical flow algorithms, which calculate the movement of pixels in the $XY$ directions with image correlation, are unable to decide the motion correctly. Learning based optical flow methods also suffer from such limitation especially for the textureless region, as shown using a rectangle in Fig.~\ref{figure:figure4}(d). The reason is that depth information is lost when it is captured by a camera in real world. As a result, there are ambiguities between the camera and object motion in the scene. These ambiguities can be resolved when using a range camera, such as stereo camera \cite{Wedel2008Efficient}. Therefore, 3D optical flow obtained from stereo sequences is a true representation of the camera motion with $XYZ$ directions in the 3D real world. In order to prove our 3D optical flow is competent to serve as supervision, an accurate comparison between 2D optical flow and our synthetic one will be described in section IV-D.

\subsection{CNN-Flow Network}\label{sec:review:slam}

It has been proved that the monocular sequences can be used to calculate the 2D optical flow with the CNNs network. Therefore, we employ the CNN-Flow network structure which is the same as FlownetC \cite{Dosovitskiy2015FlowNet, Ilg2017FlowNet}. Two consecutive images stacked together are fed into the CNNs. Then, the CNNs outputs an intermediary vector OFF whose dimension is N$\times$6$\times$20, where N is determined by the image size. Let c7s1-k denote a 7$\times$7 Convolution-InstanceNorm-ReLU layer with k filters and stride 1. The input of CNN-Flow network is merged stereo images. The network consists of:

c7s2-64,c5s2-128,c5s2-256,c3s1-256,c3s2-512,c3s1-512,c3s2-512,c3s2-1024,c3s1-1024.

Then 2D optical flow map is  refined by a ¡®upconvolutional¡¯ layers, which consists of unpooling and a convolution. Note that, synthetic 2D optical flow is used as the supervision value when the CNN-Flow network is training, thus the CNN-Flow can converge correctly. Finally, the OFF is fed into FC-Fusion network as visual input.

\subsection{LSTM-IMU Network}\label{sec:review:slam}

In recent years, the deep learning network has made initial attempts in IMU data processing. For example in VINET, a LSTM-style network is trained to learn the relative pose from the previous frame and subsequent inertial measurements to the next frame. However, the LSTM network training is usually difficult to converge because of the noises and changes of IMU data. In most traditional VIO systems, the IMU motion is calculated as a time-varying model, and the state of IMU is constantly updated. Therefore, this paper proposes a learning method considering the state of IMU.

\subsubsection{Preintegrated Network}\label{sec:review:slam}

LSTM is a time recurrent neural network which is potential for obtaining poses from IMU sequences. In this paper, IMU data and state $S$, are directly fed into LSTM. The preintegrated network is a 2-layers LSTM network and each of them has 6 hidden states. And the dropout probability is 0.5 in each layer of LSTM. Their input dimension is N$\times$12, where N is the number of IMU frames between two frames of images. Its output is the IMU-se3 with 6 dimensions (e.g. 3 translation and 3 rotation). Note that, Stereo-se3 is employed to constraint IMU-se3.

\subsubsection{Status Update Module}\label{sec:review:slam}

As mention above, our network inputs the state of the IMU, which aims to update the IMU status continuously. To accomplish this, the status error of the IMU is a feedback which is the difference of the motion between the IMU preintegrated and FC-Fusion networks. The status of IMU is updated as  follows\cite{Mur2016VIORB, Forster2015On}:
\begin{equation}
\begin{aligned}
\tilde{S}_{t-1}&=\mathop{\arg\min}_{\tilde{S}_{t-1}}({\rm VIO-}se3,{\rm IMU-}se3)\\
&=\mathop{\arg\min}_{\tilde{S}_{t-1}}(\rho([e_\mathbf{R}^Te_\mathbf{p}^T]\Sigma_I[e_\mathbf{R}^Te_\mathbf{p}^T]^T))
\end{aligned}
\end{equation}
\begin{equation}
e_\mathbf{R} = {\rm Log}((\Delta \mathbf{R}_{LSTM}(\tilde{S}_{t-1}))^T\Delta \mathbf{R}_{VIO})
\end{equation}
\begin{equation}
\begin{split}
e_\mathbf{p} = \Delta \mathbf{p}_{VIO}-\Delta \mathbf{p}_{LSTM}(\tilde{S}_{t-1})
\end{split}
\end{equation}

where $\tilde{S}_{t-1}$ is the best estimate of IMU status at time $t-1$. $\rho$ is the Huber robust cost function, $\Sigma_I$ is the information matrix of the pre-integration. $e_\mathbf{R}$ and $e_\mathbf{p}$ are the error of rotation and translation between IMU-se3 and VIO-se3. $\Delta \mathbf{R}_{LSTM}$ and $\Delta \mathbf{R}_{VIO}$ are the rotations calculated by LSTM-IMU network and FC-Fusion network from time $t-1$ to $t$. $\Delta \mathbf{p}_{LSTM}$ and $\Delta \mathbf{p}_{VIO}$ are the translations by LSTM-IMU network and FC-Fusion network from time $t-1$ to $t$.

Thereby it updates the state of the IMU every time when it receives the feedback from FC-Fusion, which is highly similar to IMU status updating in traditional tightly-coupled VIO methods.

\subsection{FC-Fusion Network}\label{sec:review:slam}

The combination of OFF and IMU-se3 is fed into the FC-Fusion network, which has 5 fully connected layers. Let fA-B-R denote a fully connected ReLU layer from A to B. dk denotes a dropout and the probability of an element to be zeroed is k. The network with 5 blocks consists of:

d0.5,f6+1024x6x20-4096-R,d0.5,4096-1024-R,f1024-128-R,f128-32-R,f32-6.

FC-Fusion is similar to multisensor fusion function in the traditional method. The output of FC-Fusion is named as VIO-se3 whose dimension is 6. Therefore, the VO trajectory of a camera over a period of time is calculated by the integration of VIO-se3. Note that, Stereo-se3 is also employed to constraint VIO-se3.

\subsection{DeepVIO Loss}\label{sec:review:slam}

\subsubsection{CNN-Flow Loss}\label{sec:review:slam}

The loss for CNN-Flow is the differences between its output and synthetic 2D optical flow within all pixels, i.e.

\begin{equation}
L_{\rm Flow}=\sum_{x,y}{\rm EPE}(\mathbf{v}_{\rm 2D}(x,y),\mathbf{v}_{{\rm CNN-Flow}}(x,y))
\end{equation}

where $\vec{V}_{\rm CNN-Flow}(x,y)$ is 2D optical flow from the CNN-Flow network. $L_{\rm Flow}$ is the summation of the endpoint error (EPE), which is the standard error used for optical flow estimation \cite{Ilg2017FlowNet}.

\subsubsection{IMU Loss}\label{sec:review:slam}

The loss of the  LSTM-IMU network is shown as follows:

\begin{equation}
L_{\rm IMU}= \Sigma (\lVert \omega-\hat {\omega} \rVert +\beta \lVert \upsilon-\hat {\upsilon} \rVert)
\end{equation}

where $\omega$ and $\upsilon$ are the camera motion true value given by Stereo-se3, $\hat {\omega}$ and $\hat \upsilon$ are the camera motion estimation given by LSTM-IMU. $\beta$ is the additional scale factor for balancing the translation and quaternion elements.

\subsubsection{Fusion Loss}\label{sec:review:slam}

The loss of the FC-Fusion is shown as follows:
\begin{equation}
L_{\rm VIO}= \Sigma (\lVert \omega-\hat \omega_1 \rVert +\beta' \lVert \upsilon-\hat \upsilon_1 \rVert)
\end{equation}

Where $\omega$ and $\upsilon$ are the quaternion and translation given by stereo-se3, while $\hat \omega_1$ and $\hat \upsilon_1$ are estimations given by FC-Fusion. $\beta'$ is the additional scale factor for balancing the translation and quaternion elements.

For the entire DeepVIO network, the total loss consists of three parts as follows:

\begin{equation}
L =L_{\rm Flow}+L_{\rm IMU}+L_{\rm VIO}
\end{equation}

\section{EXPERIMENTAL RESULTS}\label{sec:review}

In this section, we evaluate our proposed DeepVIO in comparison to the state-of-the-art algorithms on both indoor and outdoor datasets followed by detailed analysis.

\subsection{Dataset}\label{sec:review:slam}

KITTI dataset: It consists of 389 pairs of stereo images and optical flow maps, 39.2 km visual ranging sequences, a Velodyne laser scanner and a GPS/IMU localization unit, sampled and synchronized at 10Hz. The odometry benchmark consists of 22 stereo sequences, saved in loss less png format. It provides 11 sequences (00-10) with ground truth trajectories for training and 11 sequences (11-21) without ground truth for evaluation. We download the pretrained PSMNet model from this link at \url{https://drive.google.com/file/d/1p4eJ2xDzvQxaqB20A_MmSP9-KORBX1pZ/view}.

EuRoC dataset: It  contains 11 sequences recorded from a micro aerial vehicle (MAV), flying around two different rooms and an industrial environment.  The dataset provides synchronized global shutter WVGA stereo images at 20Hz with IMU measurements at 200Hz and trajectory ground truth. The pretrained model of PSMNet is trained by the rectified frames. In detail, it selects 200 training rectified stereo image pairs with the ground truth disparities obtained using an EuRoC tools (\url{https://github.com/ethz-asl/volumetric_mapping}). Image size is 752$\times$480. The whole training data is further divided  into a training set (80\%) and a validation set (20\%).

\subsection{Network Training}\label{sec:review:slam}

For the KITTI dataset, the image is rectified and resized to 640$\times$192 and IMU input data is 12$\times$1. In the model parameters, the batch size is 32. The epoch is 200 and the optimization function is Adam. In our experiment, sequences 00-08 are used for training and 09-10 are used for testing. Note that sequence 03 is abandoned since its IMU data is not available in KITTI Raw Data, In addition, 5\% of KITTI sequences 00-08 are selected as a validation set, which is the same to \cite{shamwell2018vision}.

For the EuRoC dataset, the image is 640$\times$480 and IMU input data is 12$\times$10. The batch size, epoch and optimization function are the same to KITTI. We use the same training and test sequences as \cite{wang2018deepvo}.
The computer graphics card used for training is equipped with Nvidia GeForce GTX1080 Ti with 12G memory.

\begin{figure}[!htbp]
\centering
\includegraphics[width=0.35\textwidth]{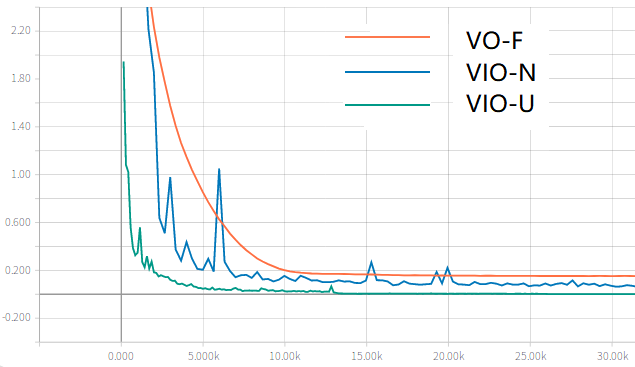}
\caption{Comparison of training losses using our models on KITTI.}
\label{figure:figure5}
\end{figure}

Fig.~\ref{figure:figure5} shows training losses for KITTI using three kinds of networks: red curve is the loss of VO-F trained only with LSTM-IMU output vector OFF, blue curve is the loss of VIO-N trained without IMU status update module, and green curve is the loss of VIO-U trained with IMU status update module. We can observe that VIO-U converges faster and has lower errors compared to others. The quantification will be shown in the section IV.E.

\subsection{Comparison of the Trajectories}\label{sec:review:slam}

We compare the performance of DeepVIO with other state-of-the-art learning-based and traditional VO or VIO baselines that are DeepVO, VIOlearner, ORB-SLAM-M (without loop closure) and VINS.

Table I shows the results in KITTI test. In order to facilitate comparison, we follow the metric provided by \cite{shamwell2018vision}, where $t_{rel}(\%)$ is the average translational error percentage on lengths 100 m - 800 m and $r_{rel}(^\circ)$ is the rotational error ($^\circ$/100m) on lengths 100m - 800m. It is observed that ours outperforms existing methods in terms of trajectory accuracy. In particular, the errors with traditional VINS are extremely worse than us. It is concluded that these traditional VIO methods are not good results for trajectory estimation when the original raw IMU data collected without tight synchronization and the IMU rate is too low in KITTI dataset.  A further experiment will be presented to analyze the performance under some extreme cases between a traditional method and ours in section IV.F. On the other hand, it can be seen that our prediction accuracy outperforms the VIOlearner except some cases. The reason is that VIOlearner just directly uses the raw IMU data to infer the trajectory without considering the change of IMU status. In contrast, our IMU status is corrected all the time when the update module receives the feedback from VIO-se3 and IMU-se3. Note that, our method is able to generalize well to unseen scenarios, e.g. seq 09 and seq 10.

\begin{figure}[!htbp]
\centering
\subfigure[KITTI 09]{%
\includegraphics[width=0.165\textwidth]{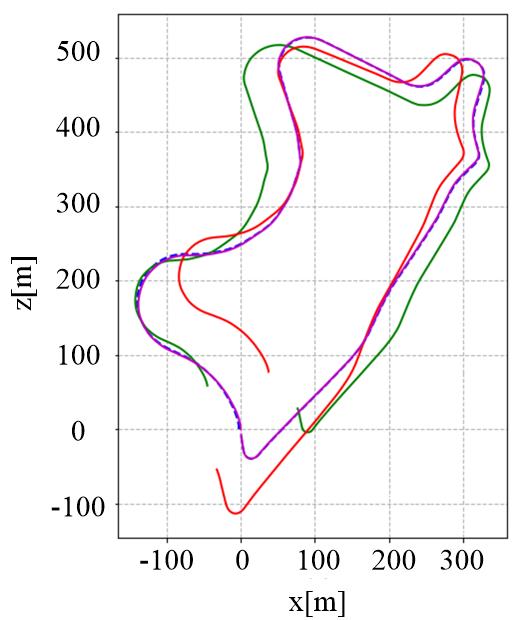}
\label{fig:subfigure1}
}
\subfigure[EuRoC MH04]{%
\includegraphics[width=0.25\textwidth]{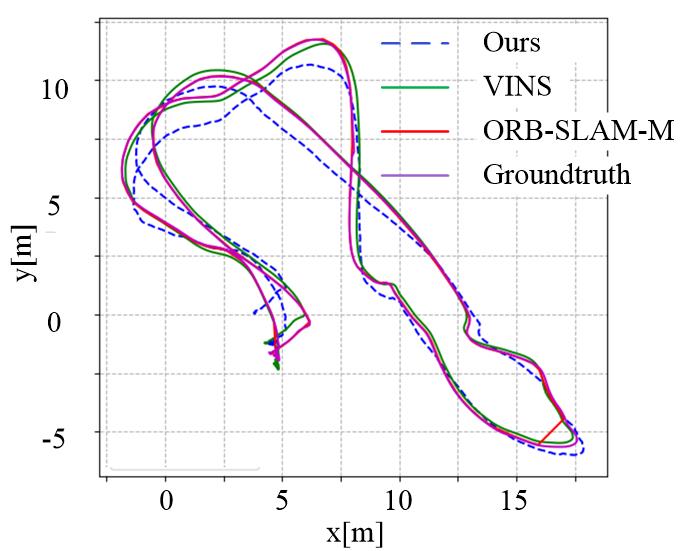}
\label{fig:subfigure2}
}
\quad
\caption{Trajectories of Ours and existing approaches testing results.}
\label{fig:figure6}
\end{figure}

We follow the test instructions \cite{Wang2017DeepVO} designed for EuRoC, which reports the results on MH04 and MH05 as shown in Table II. It shows that DeepVIO outperforms DeepVO because of combining IMU data with a camera in our cases. Our performances are slightly worse than VINS since the latter boosts via the high quality of IMU data (e.g. high frequency and good synchronization). In addition, ORB-SLAM-M also outperforms ours especially on translation accuracy. The reason is that our method lacks a module of the local bundle adjustment optimization.
The average time of DeepVIO inference are approximate 7.81ms per frame on KITTI and 3.9ms per frame on EUROC, respectively. Fig.~\ref{fig:figure6} shows the comparison of the trajectories on KITTI 09 and EuRoC MH04.

\begin{table}[!htbp]
	\centering
	\caption{COMPARISON OF TRAJECTORIES ESTIMATION PERFORMANCE WITH EXISTING APPROACHES ON KITTI DATASET.}
	\label{table1}
	\begin{tabular}{p{0.6cm}<{\centering}p{0.4cm}<{\centering}p{0.4cm}<{\centering}
    p{0.4cm}<{\centering}p{0.4cm}<{\centering}ccp{0.4cm}<{\centering}p{0.4cm}<{\centering}}
		\hline
        \multirow{2}{*}{}&\multicolumn{2}{c}{\textsl{\scriptsize{Ours}}}
        &\multicolumn{2}{c}{\textsl{\scriptsize{VIOLearner}}}&\multicolumn{2}{c}{\textsl{\scriptsize{ORB-SLAM-M}}}
        &\multicolumn{2}{c}{\textsl{\scriptsize{VINS}}}\\
		\cline{2-9}
        \scriptsize{Seq}&\multicolumn{1}{c}{$t_{rel}$}&\multicolumn{1}{c}{$r_{rel}$}&\multicolumn{1}{c}{$t_{rel}$}&\multicolumn{1}{c}{$r_{rel}$}
        &\multicolumn{1}{c}{$t_{rel}$}&\multicolumn{1}{c}{$r_{rel}$}&\multicolumn{1}{c}{$t_{rel}$}&\multicolumn{1}{c}{$r_{rel}$}\\
        \hline
        \scriptsize{00}&\scriptsize{\textbf{11.62}} &\scriptsize{\textbf{2.45}} &\scriptsize{14.27} &\scriptsize{5.29}
         &\scriptsize{25.29} &\scriptsize{7.37} &\scriptsize{18.83} &\scriptsize{2.49} \\
        \cline{2-9}
        \scriptsize{02}& \scriptsize{4.52} &\scriptsize{\textbf{1.44}} &\scriptsize{\textbf{4.07}} &\scriptsize{1.48} &\scriptsize{26.30} &\scriptsize{3.10} &\scriptsize{21.03} &\scriptsize{2.61} \\
        \cline{2-9}
        \scriptsize{05}&\scriptsize{\textbf{2.86}} &\scriptsize{2.32} &\scriptsize{3.00} &\scriptsize{\textbf{1.40}} &\scriptsize{26.01} &\scriptsize{10.62} &\scriptsize{21.90} &\scriptsize{2.72} \\
        \cline{2-9}
        \scriptsize{07}&\scriptsize{\textbf{2.71}} &\scriptsize{\textbf{1.66}} &\scriptsize{3.60} &\scriptsize{2.06} &\scriptsize{24.53} &\scriptsize{10.83} &\scriptsize{15.39} &\scriptsize{2.42} \\
        \cline{2-9}
        \scriptsize{08}&\scriptsize{\textbf{2.13}} &\scriptsize{\textbf{1.02}} &\scriptsize{2.93} &\scriptsize{1.32} &\scriptsize{32.40} &\scriptsize{12.13} &\scriptsize{32.66} &\scriptsize{3.09} \\
        \hline
        \scriptsize{09}&\scriptsize{\textbf{1.38}} &\scriptsize{1.12} &\scriptsize{1.51} &\scriptsize{\textbf{0.90}} &\scriptsize{45.52} &\scriptsize{3.10} &\scriptsize{41.47} &\scriptsize{2.41} \\
        \cline{2-9}
        \scriptsize{10}&\scriptsize{\textbf{0.85}} &\scriptsize{\textbf{1.03}} &\scriptsize{2.04} &\scriptsize{1.37} &\scriptsize{6.39} &\scriptsize{3.20} &\scriptsize{20.35} &\scriptsize{2.73} \\
        \hline
        \scriptsize{Average}&\scriptsize{\textbf{3.72}} &\scriptsize{\textbf{1.58}} &\scriptsize{4.49} &\scriptsize{1.97} &\scriptsize{25.72} &\scriptsize{7.19} &\scriptsize{21.61} &\scriptsize{2.64} \\
        \hline
	\end{tabular}
\end{table}


\begin{table}[!htbp]
	\centering
	\caption{COMPARISON OF TRAJECTORIES ESTIMATION PERFORMANCE WITH EXISTING APPROACHES ON EUROC DATASET.}
	\label{table2}
    \begin{tabular}{p{0.6cm}<{\centering}p{0.4cm}<{\centering}p{0.4cm}<{\centering}
    p{0.4cm}<{\centering}p{0.4cm}<{\centering}ccp{0.4cm}<{\centering}p{0.4cm}<{\centering}}
		\hline
        \multirow{2}{*}{}&\multicolumn{2}{c}{\textsl{\scriptsize{Ours}}}
        &\multicolumn{2}{c}{\textsl{\scriptsize{DeepVO}}}&\multicolumn{2}{c}{\textsl{\scriptsize{ORB-SLAM-M}}}
        &\multicolumn{2}{c}{\textsl{\scriptsize{VINS}}}\\
		\cline{2-9}
        \scriptsize{Seq}&{\centering}{$t_{rel}$}&{\centering}{$r_{rel}$}&{\centering}{$t_{rel}$}&{\centering}{$r_{rel}$}&{\centering}{$t_{rel}$}&{\centering}{$r_{rel}$}&{\centering}{$t_{rel}$}&{\centering}{$r_{rel}$}\\
        \hline
        \scriptsize{MH04}&\scriptsize{0.69} &\scriptsize{\textbf{0.76}} &\scriptsize{{$\approx$}5}
        &\scriptsize{$\approx$30} &\scriptsize{\textbf{0.06}} &\scriptsize{1.76} &\scriptsize{0.34} &\scriptsize{0.87} \\
        \hline
        \scriptsize{MH05}&\scriptsize{0.52} &\scriptsize{0.81} &\scriptsize{$\approx$6} &\scriptsize{$\approx$35}
         &\scriptsize{\textbf{0.05}} &\scriptsize{1.56} &\scriptsize{0.29} &\scriptsize{\textbf{0.69}} \\
        \hline
        \scriptsize{Average}&\scriptsize{0.61} &\scriptsize{0.79} &\scriptsize{$\approx$5.5} &\scriptsize{$\approx$32.5}
         &\scriptsize{\textbf{0.06}} &\scriptsize{1.66} &\scriptsize{0.32} &\scriptsize{\textbf{0.78}}\\
        \hline
	\end{tabular}
\end{table}

\subsection{Optical Flow Error}\label{sec:review:slam}

\begin{figure}[!htbp]
\centering
\subfigure[]{%
\includegraphics[width=0.2\textwidth]{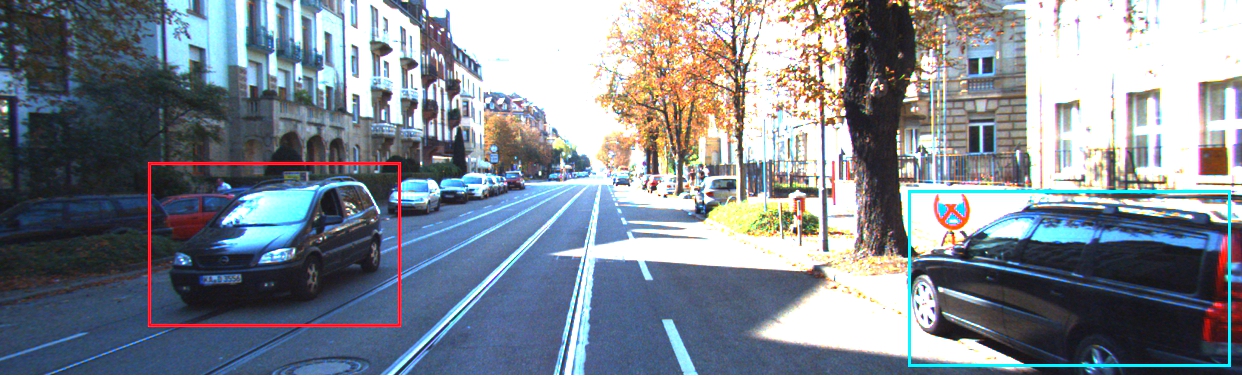}
\label{fig:subfigure1}
}
\subfigure[]{%
\includegraphics[width=0.2\textwidth]{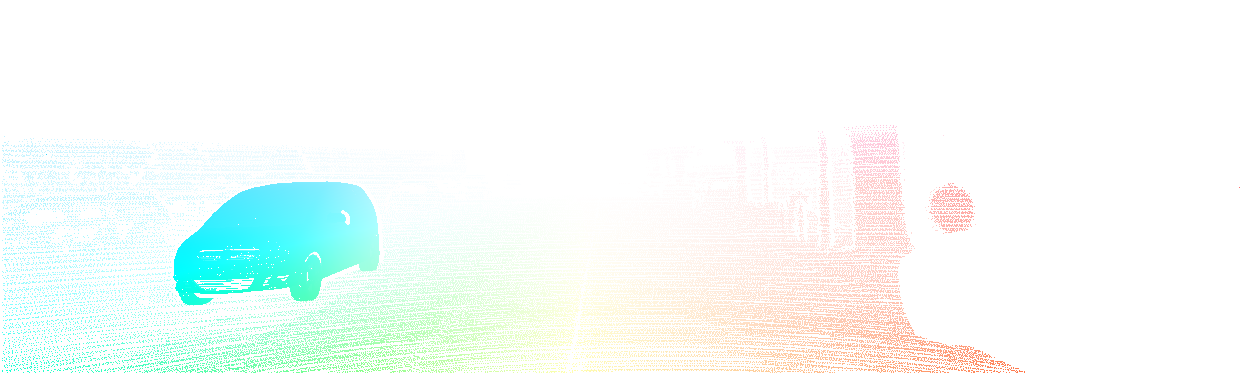}
\label{fig:subfigure2}
}
\quad
\subfigure[]{%
\includegraphics[width=0.2\textwidth]{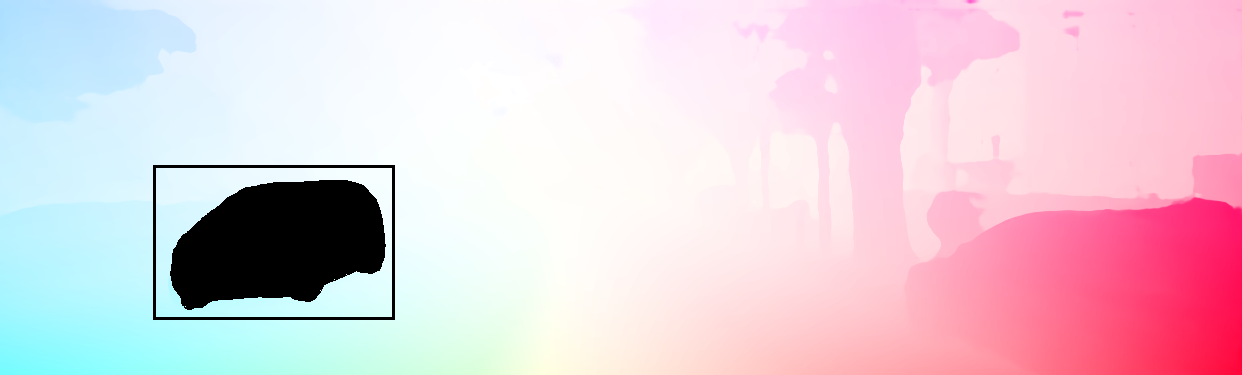}
\label{fig:subfigure3}
}
\subfigure[]{%
\includegraphics[width=0.2\textwidth]{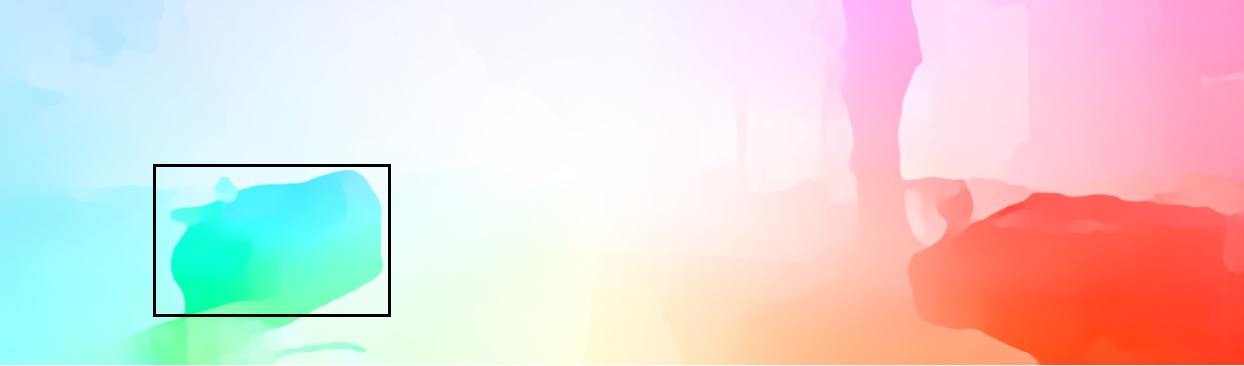}
\label{fig:subfigure4}
}
\quad
\subfigure[]{%
\includegraphics[width=0.2\textwidth]{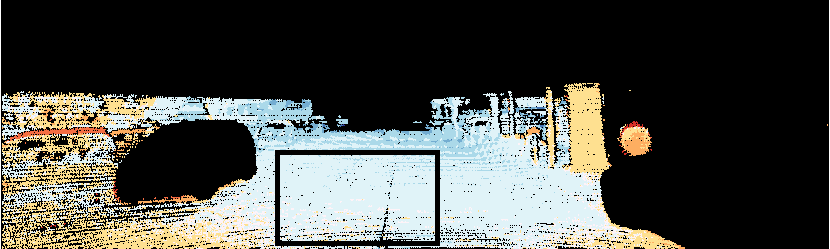}
\label{fig:subfigure3}
}
\subfigure[]{%
\includegraphics[width=0.2\textwidth]{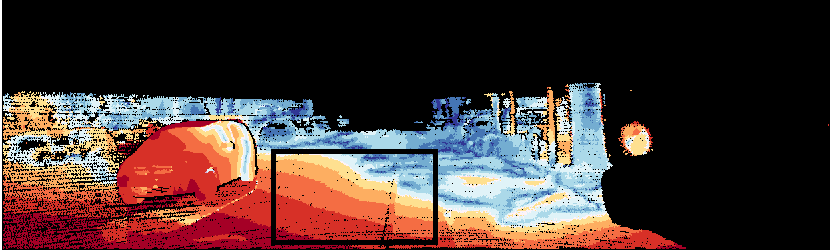}
\label{fig:subfigure4}
}
\quad
\caption[]{Comparison of  synthetic 2D optical flow and normal 2D optical flow. (a) is the left image, (b)  is its ground truth of 2D optical flow from KITTI, (c) is the synthetic 2D optical flow, (d) is the 2D optical flow obtained from Flownet2 and optical flow errors of them, e.g., (e) and (f). Red rectangle is a dynamic object and the black one is textureless road.}
\label{figure:figure7}
\end{figure}

In order to show our 3D optical flow has the potential for supervising DeepVIO. As described in section III.A, we can calculate the synthetic 2D optical flow from 3D one (see Fig.~\ref{figure:figure7}(c)) according to Eq.(7). Meanwhile, we also calculate 2D optical flow via the original FlownetC network, as shown in Fig.~\ref{figure:figure7}(d). Then, both of them are compared to the ground truth (see Fig.~\ref{figure:figure7}(b) ) provided by KITTI, and the results are visualized in Fig.~\ref{figure:figure7}(e) and (f). It is worth to mention that Flownet2 suffer from aperture ambiguity especially in the textureless region, as marked using blue rectangle in Fig.~\ref{figure:figure7}(e) and (f).  Another advantage is that more dense 2D optical flow is synthesized compared to the spare one provided by raw lidar data, which is shown as invalid region (e.g. pixel white) in the Fig.~\ref{figure:figure7}(b).

To quantify the results, we calculate the average errors of 2D optical flow on KITTI dataset in 50 pairs with non-dynamic targets and 20 pairs with dynamic targets, as shown in Table III. It can be seen that in the scene without dynamic target, the error value of 3D optical flow outperforms Flownet2 which is widely used in learning based VO methods. Compared to Flownet2 with dynamic target (see Fig.~\ref{figure:figure7} (d)), the impact of dynamic targets is eliminated when calculating the motion between two point clouds. Thus, the region of the dynamic target can be removed and the rest of them are used to train our highly accurate DeepVIO. For instance, Fig.~\ref{figure:figure7}(a) shows a black car on the left side is moving forward, and the rectangle of Fig.~\ref{figure:figure7} (c) shows our 2D optical flow marks the dynamic region as 0.

\begin{table}[!htbp]
	\centering
	\caption{Errors (in pixels) of 2D optical flow On KITTI dataset}
    \resizebox{0.4\textwidth}{6mm}{
	\label{table1}
	\begin{tabular}{ccc}
		\hline
		\scriptsize{KITTI Dataset}&\scriptsize{3D optical flow}&\scriptsize{Flownet2}\\
		\hline
		\scriptsize{Non-dynamic targets}&\scriptsize{\textbf{4.12}}&\scriptsize{5.63}\\
		\hline
        \scriptsize{Dynamic targets}&\scriptsize{\textbf{6.31}}&\scriptsize{13.62}\\
        \hline
	\end{tabular}}
\end{table}

\subsection{Comparison of DeepVIO Network}\label{sec:review:slam}

Here, we quantify the performances of the three networks (e.g. VO-F, VIO-N and VIO-U ) on KITTI sequence 10.

\begin{figure}[!htbp]
\centering
\subfigure[Translation error]{%
\includegraphics[width=0.23\textwidth]{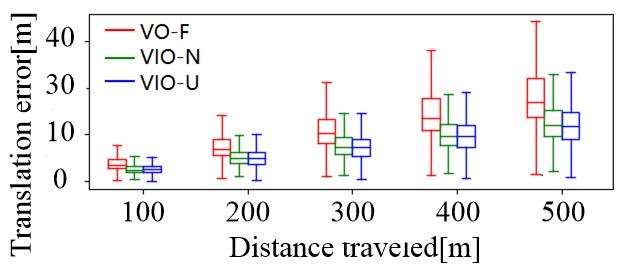}
\label{fig:subfigure1}
}
\subfigure[Rotation error]{%
\includegraphics[width=0.23\textwidth]{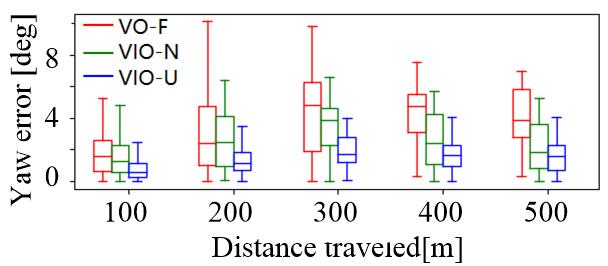}
\label{fig:subfigure2}
}
\quad
\caption{Translation and orientation errors on the KITTI 10.}
\label{figure:figure8}
\end{figure}

In Fig.~\ref{figure:figure8}, the error of the VO-F is the largest since only visual feature is fed into CNNs. The error of VIO-N is also worse than VIO-U since the former is significantly affected by the changing status in IMU data. Especially, the faster the camera rotates, the more serious the error is. It can be concluded that our IMU update module improves the location accuracy even when it is rotating rapidly, which is similar to the traditional VIO method.

\subsection{Comparison with Traditional VIO Methods}\label{sec:review:slam}

We conduct four comparative experiments on VIO calibration, sensor synchronization, camera and IMU dropped frames. Table IV indicates that DeepVIO has a better accuracy to these tests compared with the traditional VINS. More specifically, we follow VINET\cite{Clark2017VINet} to add camera-IMU calibration errors and trained DeepVIO using the mis-calibrated data (e.g. Mis-c:10$^\circ$ means 10$^\circ$ added into the rotation matrix). It shows that our method is robust to calibration errors, while VINS always has lower accuracy due to the inaccurate calibration parameters. In terms of synchronization, we randomly add 20ms to the IMU time-stamp (e.g. Unsyn:20ms).  As result, VINS seemed to be confused by the unsynchronized IMU data with a lower accuracy compared to us. The fact is that our VIO system degenerates to VO system when encountering these two situations. As mentioned above, VINS degenerated seriously when the IMU frame rate is low (e.g. IMU-D:90\% means dropped rate is 90\%). However, in DeepVIO, vision stream pays a greater role in the trajectory inference once the IMU data is missing. Finally, we random drop some frames from image sequences (e.g. Cam-D:50\% means dropped rate is 50\%) and show that VINS also had a worse performance in roation. While our IMU stream still work well to compensate the trajectory estimation with less image alignments.


\begin{table}[htbp]
	\centering
	\caption{TABLE IV. Robustness of the DeepVIO on SEVERAL Challenging Tests}
    {
	\label{table4}
	\begin{tabular}{{m{0.6cm}m{0.5cm}m{0.5cm}m{0.5cm}m{0.5cm}m{0.5cm}m{0.5cm}m{0.5cm}m{0.5cm}}}
		\hline
        \multirow{3}{*}{}&\multicolumn{2}{m{1cm}<{\centering}}{\scriptsize{Mis-c:}}
        &\multicolumn{2}{m{1cm}<{\centering}}{\scriptsize{Unsyn:}}&\multicolumn{2}{m{1cm}<{\centering}}{\scriptsize{IMU-D:}}
        &\multicolumn{2}{m{1cm}<{\centering}}{\scriptsize{Cam-D:}}\\
        &\multicolumn{2}{m{1cm}<{\centering}}{\scriptsize{${10^ \circ}$}}
        &\multicolumn{2}{m{1cm}<{\centering}}{\scriptsize{20ms}}&\multicolumn{2}{m{1cm}<{\centering}}{\scriptsize{90\%}}
        &\multicolumn{2}{m{1cm}<{\centering}}{\scriptsize{50\%}}\\
        \cline{2-9}
         \scriptsize{MH05}&{$t_{rel}$}&{$r_{rel}$}&{$t_{rel}$}&{$r_{rel}$}&{$t_{rel}$}&{$r_{rel}$}&{$t_{rel}$}&{$r_{rel}$}\\
        \hline
        \scriptsize{Ours}&\scriptsize{\textbf{0.77}} &\scriptsize{\textbf{1.11}} &\scriptsize{\textbf{0.59}} &\scriptsize{\textbf{1.02}}  &\scriptsize{\textbf{0.75}} &\scriptsize{\textbf{0.93}} &\scriptsize{0.89} &\scriptsize{\textbf{1.06}}\\
        \hline
        \scriptsize{VINS}&\scriptsize{1.38} &\scriptsize{2.88} &\scriptsize{0.65} &\scriptsize{2.93} &\scriptsize{0.88} &\scriptsize{2.94} &\scriptsize{\textbf{0.37}} &\scriptsize{3.02}  \\
        \hline
	\end{tabular}}
\end{table}

\section{CONCLUSIONS}\label{sec:conc}

This paper provided an DeepVIO to estimate the absolute trajectory of a visual-inertial sensor from stereo sequences in an self-supervised end-to-end strategy. It has been proved that the 3D geometric constraints containing 3D optical flow and 6-DoF pose play a very important role in optimizing the CNN-Flow, LSTM-IMU and FC-Fusion network. It is worth to mention that our 3D optical flow was able to eliminate the impacts of dynamic targets and textureless in the scene. Furthermore, the IMU status update scheme reduced the noise from IMU device so that it improved the LSTM-IMU inference accuracy. The experiments in KITTI and EuRoC indicated that our DeepVIO outperforms other state-of-the-art learning-based VO or VIO system in terms of pose accuracy, and generalizes well to scenarios completely unseen. Compared to traditional VIO system, DeepVIO did not need the tightly intrinsic and extrinsic parameters of the camera and IMU, which were inconvenient to obtain in VIO calibration, and had an acceptable VIO result even with poor data, such as unsynchronized and missing ones.

One limitation is that drift problem still exists in our system since the absence of place recognition and relocalization pipelines. In future, we will plan to extend its ability of loop-closure detection and global relocalization similar to traditional visual SLAM.

\bibliographystyle{unsrt}
\bibliography{ref}

\end{document}